\documentclass[letterpaper]{article} 
\usepackage{aaai25}  
\usepackage{times}  
\usepackage{helvet}  
\usepackage{courier}  
\usepackage[hyphens]{url}  
\usepackage{graphicx} 
\usepackage{natbib}  
\usepackage{caption} 
\usepackage{subcaption}
\usepackage{algorithm}
\usepackage{algorithmic}
\usepackage{amssymb}
\usepackage{tikz}
\usepackage{relsize}
\usepackage{float}

\newcommand{\tikzcmark}{%
\tikz[scale=0.23] {
    \draw[line width=0.7,line cap=round] (0.25,0) to [bend left=10] (1,1);
    \draw[line width=0.8,line cap=round] (0,0.35) to [bend right=1] (0.23,0);
}}

\usepackage{newfloat}
\usepackage{listings}
\DeclareCaptionStyle{ruled}{labelfont=normalfont,labelsep=colon,strut=off} 
\floatstyle{ruled}
\newfloat{listing}{tb}{lst}{}
\floatname{listing}{Listing}
\usepackage{booktabs}
\usepackage{placeins} 
\usepackage{afterpage}
\usepackage{hyperref}

\title{Multi-Granularity Video Object Segmentation}
\author {
    Sangbeom Lim\textsuperscript{\rm 1}\equalcontrib,
    Seongchan Kim\textsuperscript{\rm 1}\equalcontrib,
    Seungjun An\textsuperscript{\rm 3}\equalcontrib,
    Seokju Cho\textsuperscript{\rm 2},\\
    Paul Hongsuck Seo\textsuperscript{\rm 1}\correspond,
    Seungryong Kim\textsuperscript{\rm 2}\correspond
}
\affiliations {
    \textsuperscript{\rm 1}Korea University\qquad
    \textsuperscript{\rm 2}KAIST\qquad
    \textsuperscript{\rm 3}Samsung Electronics\\
    \{limsbeom, 2020320120, phseo\}@korea.ac.kr\textsuperscript{\rm 1}, \{seokju.cho, seungryong.kim\}@kaist.ac.kr\textsuperscript{\rm 2}, sjun.an@samsung.com\textsuperscript{\rm 3}
}

\makeatletter
\newcommand{\setlabel}[1]{\edef\@currentlabel{#1}\label}
\makeatother

\usepackage{bibentry}

\begin{document}

\twocolumn[{%
\renewcommand\twocolumn[1][]{#1}%
\maketitle
\vspace{-35pt}
\begin{center}
    \centering
    \captionsetup{type=figure}
    \includegraphics[width=0.95\textwidth]{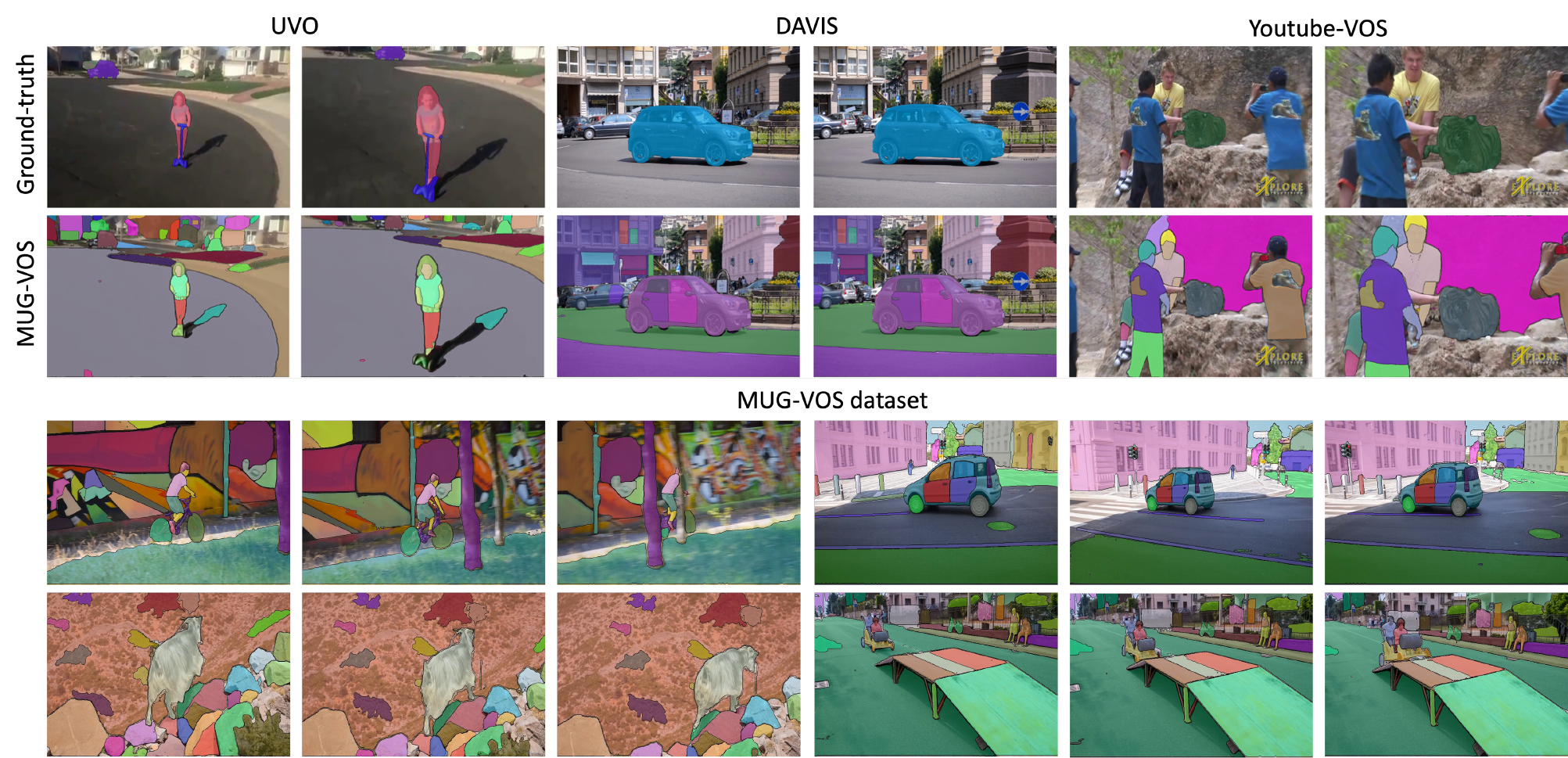}
    \caption{\textbf{Comparison of granularities of video segmentation datasets:} Visualization of \textbf{(top)} MUG-VOS masks annotated by our data collection pipeline and ground-truth masks of Youtube-VOS~\cite{xu2018youtube}, DAVIS~\cite{pont20172017}, and UVO~\cite{wang2021unidentified} data and \textbf{(bottom)} MUG-VOS dataset. MUG-VOS masks include various types and granularities of objects, parts, stuff, and backgrounds, even those not covered by existing datasets.}
    \label{fig:dataset_comparison}
\end{center}%
}]

\begin{abstract}
Current benchmarks for video segmentation are limited to annotating only salient objects (i.e., foreground instances). Despite their impressive architectural designs, previous works trained on these benchmarks have struggled to adapt to real-world scenarios. Thus, developing a new video segmentation dataset aimed at tracking multi-granularity segmentation target in the video scene is necessary. In this work, we aim to generate multi-granularity video segmentation dataset that is annotated for both salient and non-salient masks.
To achieve this, we propose a large-scale, densely annotated multi-granularity video object segmentation (\textbf{MUG-VOS}) dataset that includes various types and granularities of mask annotations. We automatically collected a training set that assists in tracking both salient and non-salient objects, and we also curated a human-annotated test set for reliable evaluation.
In addition, we present memory-based mask propagation model (MMPM), trained and evaluated on MUG-VOS dataset, which leads to the best performance among the existing video object segmentation methods and Segment SAM-based video segmentation methods.
Project page is available at \href{https://cvlab-kaist.github.io/MUG-VOS}{\texttt{https://cvlab-kaist.github.io/MUG-VOS}}.
\end{abstract}
\section{Introduction}
Video segmentation has been one of fundamental tasks in computer vision, aimed at predicting and tracking the masks corresponding to specified targets in video data~\cite{yang2019video, kim2020vps, wang2021unidentified}. Video segmentation showed great performance on identifying object and consistently tracking the object, however has shown poor performance on unknown class that has not trained on supervised stage. To develop a model capable of detecting and segmenting a wide range of categories, it is necessary to construct a larger dataset that includes as many classes as possible. However, this approach is cost-inefficient and challenging, making it difficult to create a model that can handle various types of object masks.

To alleviate this burden, several video segmentation tasks such as video object segmentation (VOS)~\cite{caelles2017one} or interactive video object segmentation~\cite{benard2017interactive} provide additional references for the target by giving segmentation mask at the first frame or giving iterative refinement to clarify it.
Unfortunately, these tasks also encounter performance degradation when they aim to segment non-salient objects, as they are only trained on specific targets, such as objects and stuff, according to the training dataset.
Their utility in real-world applications often struggles in certain cases, such as interactive video editing and open-world video understanding, which require to return valid segmentation mask tracks for any given segmentation prompt.

Regardless of the model's architecture, segmenting anything in a video scene requires extensive data. SA-1B~\cite{kirillov2023segment}, an image segmentation dataset for the Segment Anything task, averages 100 masks per image. In contrast, most video segmentation datasets~\cite{athar2023burst} have far fewer mask tracks per video and rely on costly human annotation, which, while reliable, is resource-intensive.


In this work, we introduce \textbf{Mu}lti-\textbf{G}ranularity \textbf{V}ideo \textbf{O}bject \textbf{S}egmentation (\textbf{MUG-VOS}) dataset to extend the success of segment anything to the video domain. Our goal is to provide annotations of varying granularity for target masks, covering a range of objects, parts, and backgrounds. 
MUG-VOS fundamentally differs from previous video segmentation datasets~\cite{xu2018youtube, pont20172017, wang2021unidentified} by addressing limitations of traditional Video Instance Segmentation and Video Panoptic Segmentation, which often relies on a closed vocabulary, such as the 40 categories in YouTube-VIS~\cite{yang2019video} or the 124 classes in VIPSeg~\cite{miao2022large}. While Open-vocab Video Segmentation~\cite{wang2023towards} manages unlimited classes via natural language descriptions, it typically focuses on salient objects and often miss background elements or partial objects.

To build a large-scale video segmentation dataset capable of segmenting various targets, we utilized SAM, which can segment anything in images. We propose a SAM-based data collection pipeline that leverages a simple IoU-based tracking method to generate diverse training data. This resulted in a new dataset with 77k video clips and 47M masks, designed to train and evaluate segmentation and tracking at multiple granularities masks. MUG-VOS includes diverse mask tracks—salient, non-salient, and partial objects—enabling training and evaluation of models to predict previously unmanageable masks in video segmentation.
Fig.~\ref{fig:dataset_comparison} clearly shows the difference between the MUG-VOS dataset and existing video segmentation datasets. 


On the other hand, recent attempts to adapt SAM for video segmentation~\cite{zhou2024sam} have faced challenges by using SAM primarily as an image segmentation tool. For example, SAM-PT~\cite{rajivc2023segment} tracks points from the first frame to propagate masks, while DEVA~\cite{cheng2023tracking} uses SAM to generate candidate masks for each frame, while XMem~\cite{cheng2022xmem} propagates these masks by matching them with IoU.

SAM's strength lies in its ability to generate masks of varying granularity from prompts. If this capabilities extend to the video domain, it could be highly valuable for real-world applications like interactive video editing~\cite{lee2023shape, zhang2024avid, lee2023one} and generation, where traditional methods fall short.


We also propose a Memory-based Mask Propagation Model (MMPM) as a baseline for MUG-VOS . For MMPM, we utilize the pre-trained SAM encoder to leverage the rich knowledge of SAM. Specifically, we introduce a memory module that can store information about target objects from previous outputs. This memory module allows the model to generate consistent segmentation mask tracks throughout the video. MMPM processes video frames sequentially while attending the memories that are relevant to the current frame. This module enables the SAM structure to be directly applied in the video domain. Additionally, we evaluate existing video segmentation methods and MMPM on MUG-VOS test dataset. MMPM shows best performance quantitatively and qualitatively on MUG-VOS dataset.
\section{Related work}
\subsection{Video segmentation}
The segmentation task stands out as one of the most extensively researched areas within the field of computer vision, encompassing various sub-tasks such as instance segmentation~\cite{wu2023general, zhang2023dvis, Meinhardt2023NOVISAC}, semantic segmentation~\cite{hu2020tdnet, wang2021temporal, liu2020efficient}, and panoptic segmentation~\cite{li2022videoknet, vip_deeplab, athar2023tarvis}.

Video semantic segmentation focuses on segmenting the same semantic class referring to segmentation results on different frames. This method mainly focuses on improving the temporal consistency of segmented results.
Video instance segmentation further challenges the complex problem of consistently identifying the same instance even on the deformation, blur, and existence of similar object classes.
Video panoptic segmentation~\cite{kim2020vps} further challenges to segment background stuff as previous methods have not been discovered.

Despite the remarkable performance achieved by these methods, they often struggle when confronted with unseen classes, limiting their applicability in real-world scenarios.
To address this limitation, the concept of open-world segmentation~\cite{Xu_2023_CVPR, liu2022opening} has been introduced. In this paradigm, models are designed to handle not only known classes that have been seen on the train but also unseen ones. For example, the BURST~\cite{athar2023burst} dataset contains 482 labels, which can be categorized into 78 seen classes and 404 unseen classes. Similarly, the UVO~\cite{wang2021unidentified} dataset has been developed to facilitate open-world segmentation, aiming to build class-agnostic segmentation models. These approaches prioritize the ability to segment unseen classes, albeit often at the expense of overall performance. In contrast, our model is specifically engineered to segment any class based on given input conditions, without compromising performance or being restricted by class distinctions.

\subsection{Segment anything}
SAM~\cite{kirillov2023segment} has emerged as a powerful solution for image segmentation tasks, demonstrating impressive performance. What sets SAM apart is its ability to interpret user input in various forms, including points, bounding boxes, and text. This ability to interpret any prompt format enables users to provide segmentation guidance through multiple modalities, enhancing the model's usability and flexibility.
Another key strengths of SAM lies in its iterative training approach, facilitated by the utilization of SA-1B~\cite{kirillov2023segment}, a dataset containing various granularity of masks. Through this iterative process, SAM refines its segmentation capabilities, learning from the data in SA-1B and continuously improving its performance.
This iterative training strategy plays a crucial role in enhancing SAM's effectiveness, allowing it to adapt to diverse inputs. As a result, SAM has proven to be a highly effective tool for a wide range of image segmentation applications, offering promising results and paving the way for further advancements in this field.

\subsection{Segment anything in video domain}
In recent years, a variety of approaches have emerged to extend SAM to the video level. For instance, Tracking Anything~\cite{zhu2023tracking, cheng2023tracking} and DEVA~ \cite{cheng2023tracking} fused SAM at the image level and propagated masks to future frames using a video object segmentation model. SAM-PT~\cite{rajivc2023segment} adopted a different strategy by sampling points from the mask and leveraging a point tracking model to estimate their positions in future frames. These estimated points were then utilized to generate masks using SAM. Meanwhile, SAM-PD~\cite{zhou2024sam} tackled video segmentation as a prompt denoising task. Their method involved spreading jittered and scaled bounding boxes and then selecting them based on semantic similarity.
However, these prior works employ independent methods for propagating masks. This approach introduces noise that accumulates errors as the timestep increases.
In contrast, our method takes an end-to-end approach to handle video segmentation, eliminating the need for additional steps and thus mitigating error accumulation over time.
\begin{figure*}[h!]
    \vspace{-10pt}
    \includegraphics[width=\linewidth]{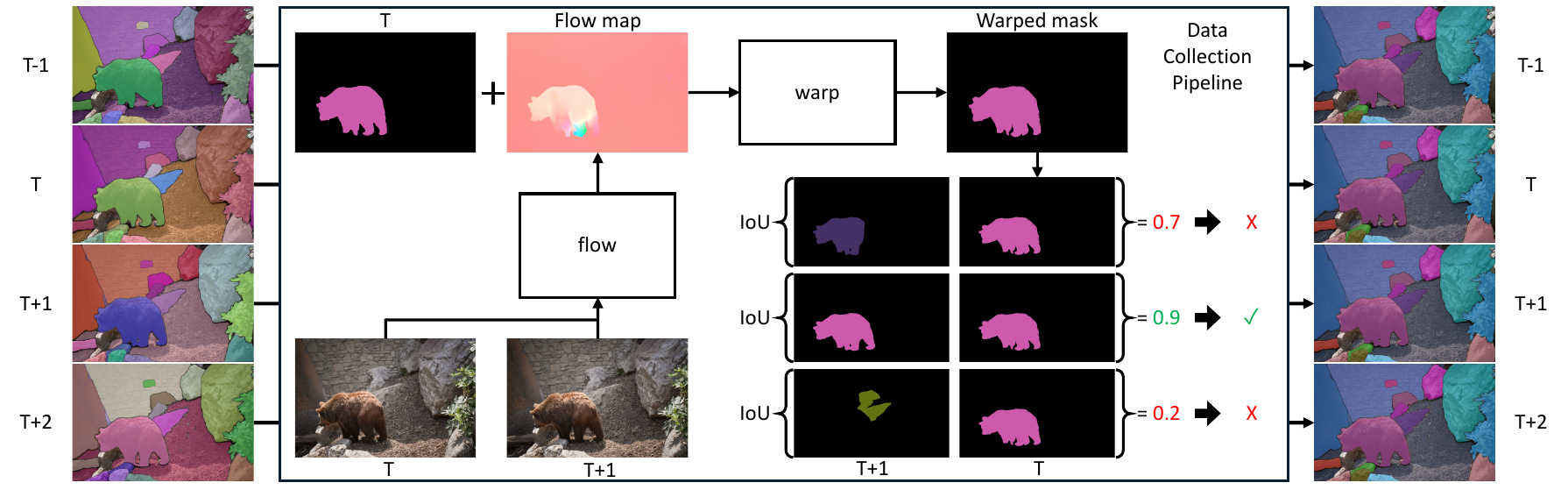}
    \vspace{-10pt}
        \caption{\textbf{MUG-VOS data collection pipeline.} We propose a data collection pipeline to generate a dataset to curate multi-granularity mask tracks completely automatically. Using SAM, we generate a large number of masks per frame and find a temporal connection through the IoU between the mask warped from the previous frame and the mask from the current frame.}
    \label{fig:engine}
    \vspace{-10pt}
\end{figure*}

\section{Dataset}
Existing video segmentation datasets~\cite{xu2018youtube, pont20172017, wang2021unidentified, Miao_2022_CVPR, athar2023burst} typically focused on salient target objects that can be defined as specific classes. 
However, there currently exists no dataset for training or evaluating a diverse granularity of masks in video segmentation task.

Despite the numerous datasets for image-level segmentation, video-level segmentation datasets are scarce. Additionally, the lack of densely annotated masks in current video datasets hinders the adaptation of SAM on video segmentation.
In order to train and evaluate MMPM, a large-scale video segmentation dataset is needed. To satisfy such conditions, Annotating a mask in a handcrafted way is accurate, but costs a lot. 
Therefore, we employ a semi-automatic data generation process involving minimal human annotators. To create a sufficiently large and diverse dataset, we developed a data collection pipeline to collect video segmentation data with varied granularity masks. Our data collection pipeline produces the segmentation data in a step-wise manner, ensuring both the quality and quantity of the masks. 

\subsection{Data collection pipeline}
We propose a data collection pipeline designed to autonomously derive video segmentation data from pre-existing video sources. Our data collection pipeline exploits SAM to generate dense masks for each frame and leverages an optical flow model to establish temporal connections between masks across consecutive frames, thereby generating pseudo-masks for video segmentation.

More specifically, for the given video \(V=\{v_t\vert t\in [1,T]\}\), where \(v_t\) is $t$-th frame image, and \(T\) is the number of frames, target masks \(M_t^i\) are defined as those obtained from the initial frame using SAM through grid point prompts. Masks from subsequent frames that are temporally consistent to the masks from the initial frame, where \(i\) is the index of the track containing the target masks. Following the initial frame, prior to obtaining the target mask in the \(t\)-th frame, candidate masks potentially serving as the target mask are predicted. 

To generate the candidate masks \(C_t^i\), the points \(P_t^i\) are sampled from the target mask \(M_{t-1}^i\). Subsequently, the points \(W_t^i\) are derived by warping the sampled points \(P_t^i\) into the \(t\)-th frame utilizing a flow map \(F_{t-1 \rightarrow t}\) predicted by a flow model.
\begin{equation}
    P_t^i = \mathrm{sample}(M_{t-1}^i) \\
\end{equation}
\begin{equation}
    W_t^i = \mathrm{warp}(P_t^i, F_{t-1\rightarrow t})
\end{equation}
\begin{equation}
    F_{t-1\rightarrow t} = \mathrm{flow}(v_{t-1}, v_t)
\end{equation}
The \(\mathrm{sample}\) function denotes the process of randomly selecting point coordinates on the mask, while the \(\mathrm{warp}\) function represents the warping of a point or a mask according to the flow map. The \(\mathrm{flow}\) function refers to the estimation of a flow map between two adjacent frames.
Candidate masks \(C_t^i\) in the \(t\)-th frame are derived by applying the point prompts acquired from the warped points \(W_t^i\) to SAM. The target mask \(M_{t-1}^i\) is further warped to the \(t\)-th frame via the flow map \(F_{t-1\rightarrow t}\) to obtain the warped mask \(\tilde{M}_{t-1}^i\). Subsequently, the Intersection over Union (IoU) between the warped mask \(\tilde{M}_{t-1}^i\) and the candidate masks \(C_t^i\) are computed, with the candidate mask exhibiting the highest IoU being designated as the target mask \(M_t^i\) in the \(t\)-th frame.
\begin{equation}
    C_t^i = \mathrm{SAM}(v_t, W_t^i)
\end{equation}
\begin{equation}
    \tilde{M}_{t-1}^i = \mathrm{warp}(M_{t-1}^i, F_{t-1\rightarrow t})
\end{equation}
\begin{equation}
\label{iou_eq}
    M_t^i = \arg\max_{C_t^i}\mathrm{IoU}(\tilde{M}_{t-1}^i, C_t^i)
\end{equation}

\paragraph{Quality assurance.}
Fully automated processes can induce significant errors in the evaluation procedure. To prevent errors from accumulating in the automated process, human annotators were tasked with manually tracking and generating masks. For efficiency, annotators were instructed to accept or reject the mask tracks produced by the data collection pipeline. If an error occurred, the annotators refined the mask at the frame level using SAM. More details can be found in the appendix.

To ensure the quality of the mask tracks in the dataset, we adopted a verification process. As human annotators conduct the tracking process, completed tracks are sent to a supervisor for approval. The supervisor reviews the annotated mask tracks alongside the original video to determine whether the tracks are satisfactory or unsatisfactory. Rejected mask tracks are sent back to the annotators for refinement.
Quality assurance process was only done on the MUG-VOS test dataset.

\subsection{MUG-VOS Dataset}
The MUG-VOS dataset was built using our data collection pipeline and, to the best of our knowledge, is larger than any other video segmentation dataset.

\paragraph{Video collection.}
In the pursuit of crafting the video segmentation dataset, a judiciously selected subset of videos from the HD-VILA-100M~\cite{xue2022hdvila} dataset was scrupulously curated for inclusion. The HD-VILA-100M dataset is devised to function as an expansive, high-resolution, and diversified video-language dataset, with the overarching goal of fostering multi-modal representation learning. This dataset encompasses a total of 3.3 M videos, characterized by their high caliber and equitable distribution across 15 categories. A subset of videos from this dataset was processed to get 77,994 video clips.

In addition, We create additional annotations for the 30 videos in the DAVIS-17\cite{pont20172017} validation set, which widely used in VOS task, for MUG-VOS test sets. We use SAM to generate average 29.6 mask tracks of varying types and granularities for each video.

\paragraph{Data generation.}
The MUG-VOS dataset facilitates training and evaluation on masks with various types and granularities, which are not covered by existing video segmentation datasets. Our data collection pipeline simplifies this process while ensuring high quality. To achieve this, we adapted our pipeline to the HD-VILA-100M and DAVIS-17 videos, resulting in a dataset that is both high quality and large in scale.


\paragraph{Data statistics.}
\begin{table*}[t]
\centering
\begin{center}
\begin{tabular}{l @{\hspace{5mm}} c @{\hspace{5mm}} c @{\hspace{5mm}} c @{\hspace{5mm}} c @{\hspace{5mm}} c}
\toprule
 & Density & Masks & Mask Tracks & Masks per Frame & Annotated Frames \\
\midrule
OVIS~\cite{qi2022occluded} & 0.186 & 296k & 5,223 & 5.8 & 51,059 \\
YT-VIS~\cite{yang2019video} & 0.167 & 132K & 4,866 & 1.69 & 79,260 \\
YT-VOS~\cite{xu2018youtube} & 0.184 & 17K & 8,614 & 1.63 & 123,265 \\
DAVIS~\cite{pont20172017} & 0.120 & 27K & 386 & 2.6 & 150 \\
UVO~\cite{wang2021unidentified} & 0.425 & 593K & 104,898 & 12.3 & 58,140 \\
BURST~\cite{athar2023burst} & 0.167 & 600K & 16,089 & 3.1 & 195,713 \\
VIPSeg~\cite{Miao_2022_CVPR} & 0.979 & 926K & 38,592 & 10.9 & 3,536 \\
\midrule
\textbf{MUG-VOS Train} & 0.714 & 47M & 4.7M & 66.3 & 77,9940 \\
\textbf{MUG-VOS Test} & 0.663 & 59K & 887 & 29.6 & 1,999 \\
\bottomrule
\end{tabular}
\end{center}
\vspace{-5pt}
\caption{\textbf{Data comparison with previous video segmentation datasets.} Comparing MUG-VOS with existing video segmentation datasets in terms of statistics.}
\vspace{-10pt}
\label{Table:statistics}
\end{table*}

%


Table~\ref{Table:statistics} presents the evaluation of related datasets and benchmarks for video segmentation. The density \( D \) of the segmentation masks is notated in Table~\ref{Table:statistics} can be given by the equation~\ref{eq:density}:
\begin{equation}
\label{eq:density}
    D = \frac{\sum_{i=1}^{H} \sum_{j=1}^{W} M_{i,j}}{H \times W} \\    
\end{equation}
where as \( H \) is the height of the image, \( W \) is the width of the image, \( M_{i,j} \) is the value of the mask at pixel \((i,j)\), which is 1 if the pixel is covered by any mask and 0 otherwise.

For VOS tasks, common benchmarks include YouTube-VOS~\cite{xu2018youtube} and DAVIS~\cite{pont20172017}. Both datasets are annotated on in-the-wild videos, each lasting approximately 5-10 seconds. However, these datasets contain only a few mask tracks per video, which limits their ability to evaluate tracking performance across diverse masks.

The UVO~\cite{wang2021unidentified} dataset primarily focuses on evaluating performance on unseen mask classes during training. It provides mask annotations for a variety of salient objects that humans typically recognize as ``objects''. While UVO covers a wide range of masks from a human perspective, it does not include non-salient objects that are not easily defined by humans, which remains as a limitation.

VIPSeg~\cite{Miao_2022_CVPR} is part of the video panoptic segmentation benchmark. Video panoptic segmentation aims to predict the semantic class of every pixel in the temporal dimension. Compared to the datasets in Table~\ref{Table:statistics}, VIPSeg covers a higher density of pixels than any other dataset in Table~\ref{Table:statistics}. However, the masks in VIPSeg are much simpler compared to MUG, as the number of masks per frame is significantly lower than in MUG.

BURST~\cite{athar2023burst} is a universal dataset designed to cover multiple segmentation tasks, including video object segmentation, video instance segmentation, and point-guided segmentation~\cite{guo2024vanishing, zulfikar2024point}.
While the aforementioned datasets primarily focus on annotating salient objects (e.g., humans, cars, animals), MUG-VOS includes a diverse range of class-agnostic masks.


\begin{figure*}[!hbtp]
    \centering
    \includegraphics[width=\linewidth]{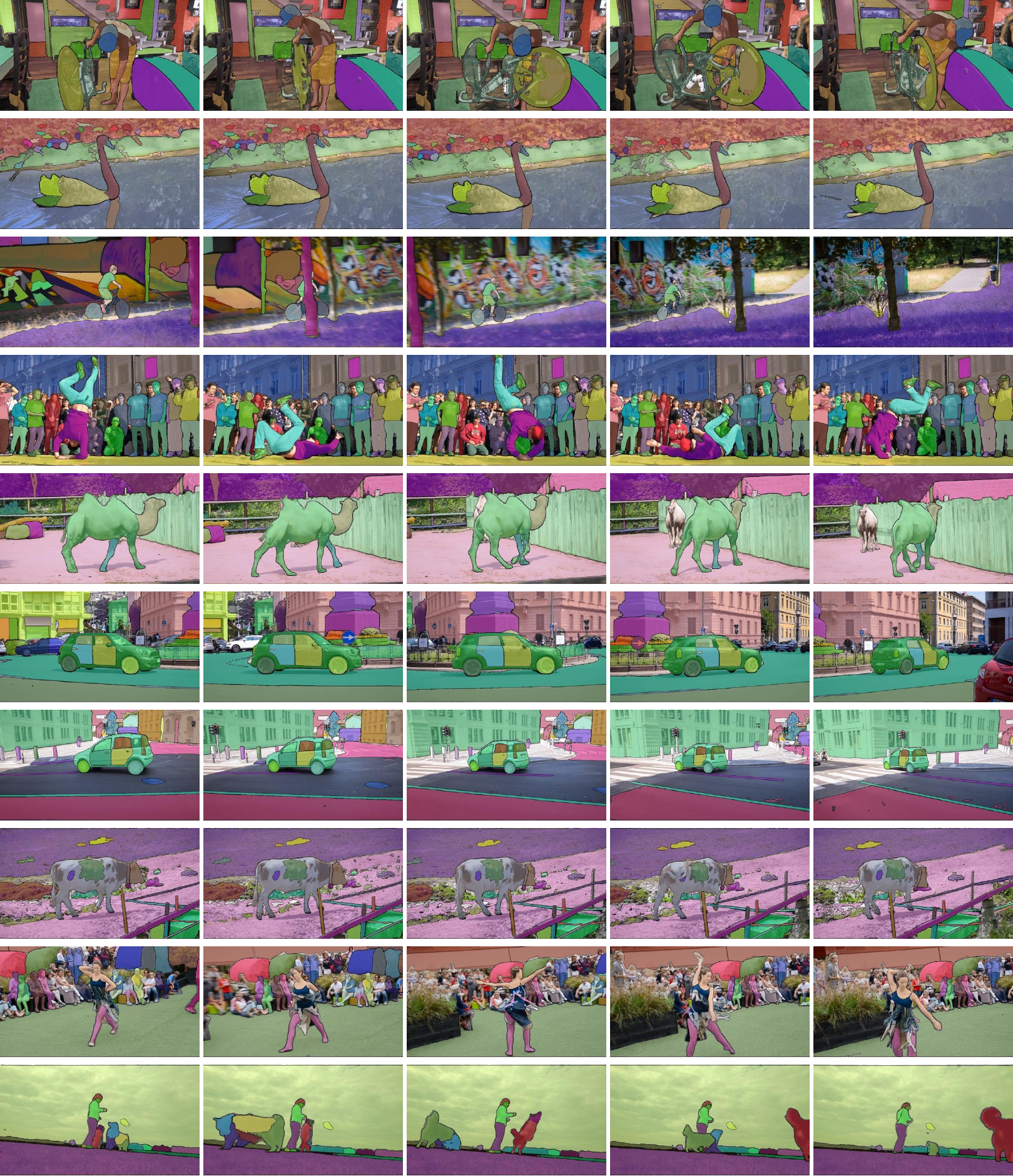}
    \caption{\textbf{MUG-VOS test dataset.} To mitigate the accumulation of errors within the automated process, annotators were directed to either approve or reject the mask tracks generated by the data collection pipeline. In instances where errors were detected, the annotators performed frame-level refinements of the masks utilizing the Segment Anything Model~\cite{kirillov2023segment}.}
    \label{fig:arxiv_mug_test_0}
\end{figure*}


Figure~\ref{fig:arxiv_mug_test_0} shows visualization of our MUG-VOS test set.

\section{Model}
Figure~\ref{fig:overview} shows the overview of MMPM. For simplicity, the figure assumes a single object, but the MMPM model is designed to handle multiple objects simultaneously. MMPM consistently tracks objects and generates segmentation masks sequentially, frame-by-frame. Thanks to the memory module, which is designed to handle occlusion, motion blur, and deformation, the model can effectively track objects even in challenging video scenes.
To build a robust video segmentation model, we incorporated two types of memory modules that function independently. For each frame, the RGB image is encoded by an image encoder, which serves as a query for the memory module. The encoded query from the image encoder accesses the memory bank to retrieve memory features that provide crucial information for generating the current mask. MMPM uses two types of memory: sequential memory and temporal memory, both of which are updated for selected frame. Frames selected for memory update are selected at regular intervals. Specifically, the temporal memory has a storage limit, $T_{max}$ , to prevent running out of memory. When the temporal memory reaches a store limit, we randomly filter out memory entries, except for those from the first and previous frames. These memory features works as a enhancement for image feature to generate high-quality masks, even in long videos.

\subsection{Temporal memory}
The temporal memory stores information about previous predictions for the same target object in the video by maintaining high-resolution features for up to $T_{max}$ recent frames. We split the memory into key and value components, which effectively retrieve relevant information. The key is encoded in the same embedding space as the query feature, while the value is encoded from a value encoder that fuses semantic and spatial information by encoding the image and binary mask together. Specifically, the key $\mathbb{R}^{C^{k}\times THW}$ exists in the same embedding space as the query feature, while the value is encoded independently.

For a selected frame, we copy the query feature from the image encoder as a key for the current frame. The generated key is then processed along with the binary mask from the mask decoder through the value encoder. The new key and value from the current frame are then appended to the memory and provide important information for future decoding processes. When the number of temporal values reaches $T_{max}$, we randomly sample values, excluding those from the first and previous frames. The first frame's value provides initial information about the target object in the VOS task and is considered to best represent the target object. The value from the last frame contains the latest information about the target object.

This dynamic filtering of the temporal memory enables the model to represent both short-term and long-term object motion effectively.

\subsection{Sequential memory}
While temporal memory focuses on high-resolution features from previous frames, sequential memory focuses on low-resolution features such as geometric, semantic, and location information. Sequential memory complements temporal memory by updating the value on selected frame.

Specifically, sequential memory $S\in\mathbb{R}^{C^{s}\times HW}$ is updated every select frame using the query from the current frame. To effectively propagate sequential information, we use the GRU~\cite{cho2014properties} method, which dynamically forgets previous information and updates with the current frame's information. The output from the value encoder serves as a candidate for updating the sequential memory. The GRU discards outdated information from the previous sequential memory and updates the sequential memory in parallel. This propagation process compensates for the lack of object information in temporal memory and stores new low-resolution information.

\begin{figure*}[h!]
    \centering
    \vspace{-5pt}
    \includegraphics[width=\linewidth]{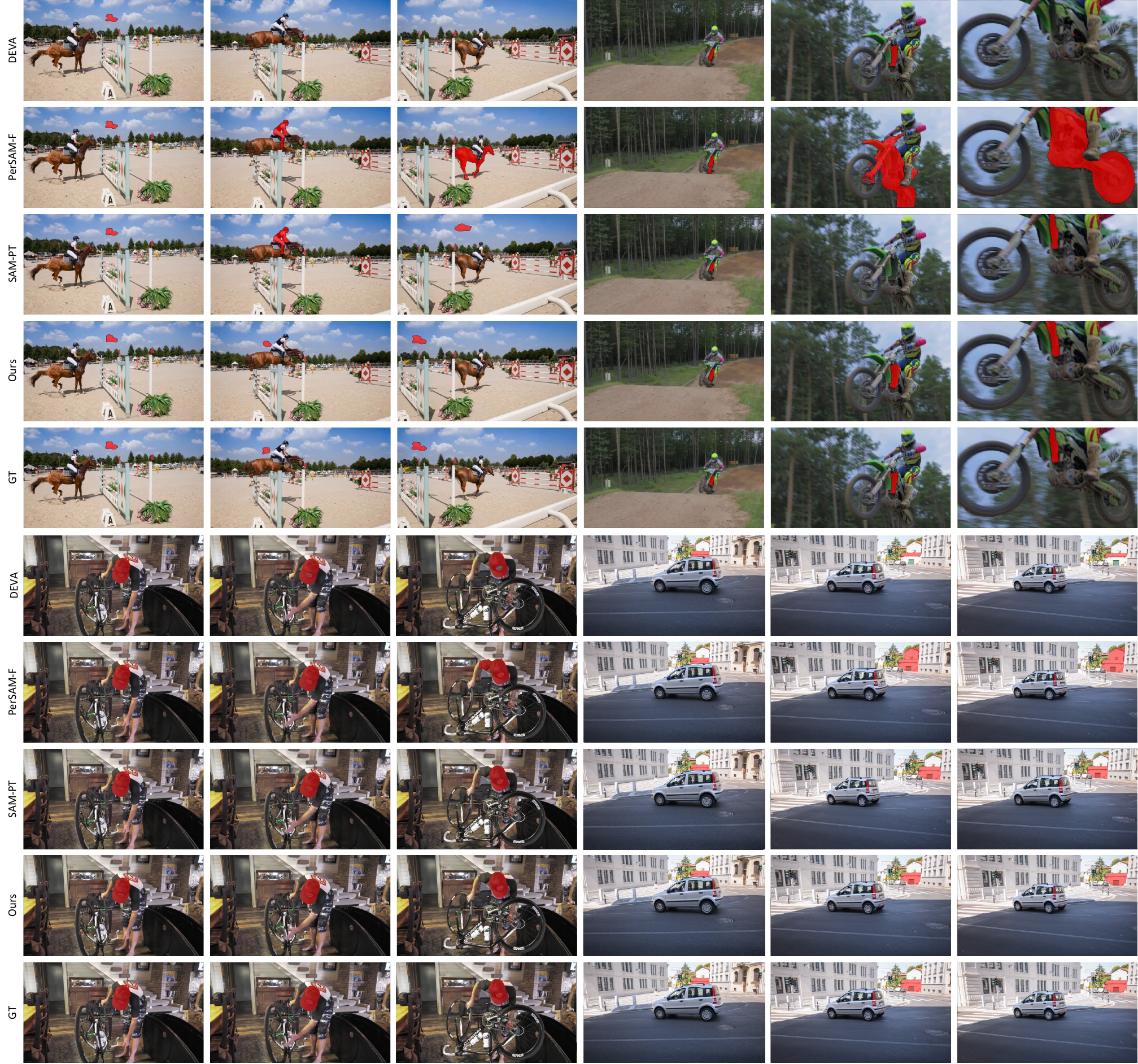}
    \vspace{-10pt}
    \caption{Qualitative comparison between MMPM, DEVA~\cite{cheng2023tracking}, PerSAM-F~\cite{zhang2023personalize}, and SAM-PT~\cite{rajivc2023segment} from MUG-VOS test set.}
    \label{fig:quality_comparision}
    \vspace{-5pt}
\end{figure*}



\begin{figure*}[t]
    \vspace{-10pt}
    \includegraphics[width=\linewidth]{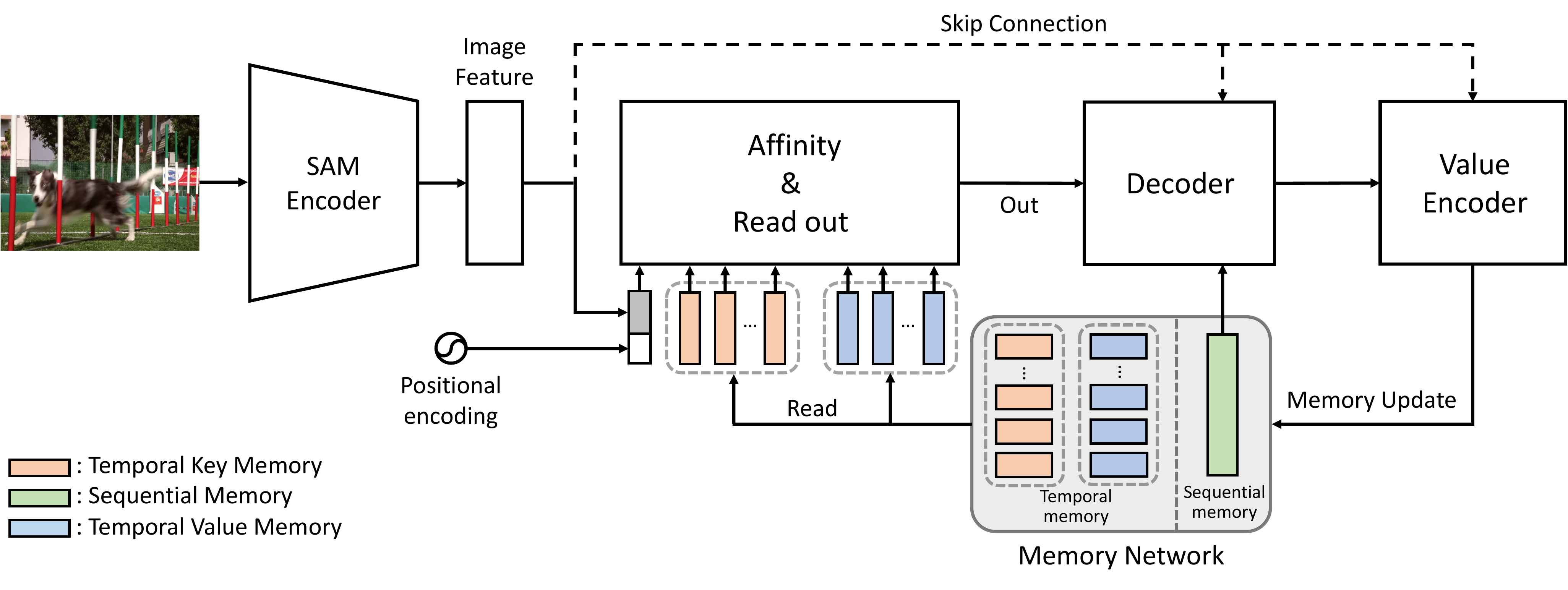}
    \vspace{-15pt}
        \caption{\textbf{MMPM overview.} We introduce the MMPM model, which generates masks based on previous results. Starting from an initial mask that indicates the target object, the MMPM model consistently tracks and segments the target throughout the entire video. Sequential memory stores low-resolution features, updated at every selected frames, while temporal memory retains high-resolution features from previous frames, capturing a variety of information gathered from multiple frames.
        }
    \label{fig:overview}
\end{figure*}
\section{Experiments}
\label{others}

\subsection{Evaluation metric}

For the evaluation metric, we use the Jaccard Index ($\mathcal{J}$), the boundary F1-score ($\mathcal{F}$), and the average of these metrics ($\mathcal{J} \& \mathcal{F}$) in the MUG-VOS dataset. In MUG-VOS, evaluation is performed independently for each individual mask, whereas in video object segmentation, evaluation is performed simultaneously across all masks in a frame.

\begin{table*}[t]
\centering 
\begin{tabular}{lcccccc}
\toprule

MUG-VOS Test  &  & SAM & $\mathcal{J}\&\mathcal{F}$ & $\mathcal{J}$  & $\mathcal{F}$\\
\midrule
\textit{zero-shot transfer} & \\
SAM (Grid 32) + IoU & & $\tikzcmark$ & 22.2 & 20.2 & 24.2 \\
SAM (Grid 64) + IoU & & $\tikzcmark$ & 26.7 & 24.5 & 28.9 \\
PerSAM~\cite{zhang2023personalize} & & $\tikzcmark$ & 33.1 & 30.8 & 35.3 \\
PerSAM-F~\cite{zhang2023personalize} & & $\tikzcmark$ & 41.5 & 39.1 & 43.9 \\
SAM-PT~\cite{rajivc2023segment} &  & $\tikzcmark$ & 78.3 & 76.2 & 80.4 \\

\midrule

XMem~\cite{cheng2022xmem} & & - & 83.0 & 86.9 & 79.1 \\
DEVA~\cite{cheng2023tracking} &  & - & 85.6 & \underline{88.2} & 82.9 \\

\midrule
MMPM & & - & \underline{86.1} & 86.0 & \underline{86.1} \\

\bottomrule
\end{tabular}
\caption{The quantitative evaluation of video object segmentation on DAVIS-2017~\cite{pont20172017} validation set. Note that all the models are trained on the MUG-VOS train set. ``SAM" correspond to an architecture that used SAM~\cite{kirillov2023segment} Encoder and Decoder.}
\vspace{-10pt}
\label{Table:visa}
\end{table*}

\subsection{MUG-VOS evaluation}
In order to evaluate the capability of multi-granularity segmentation in a video scene, we tested our models on the MUG-VOS dataset, which contains high-quality masks with a variety of granularities. To demonstrate that our MMPM model is best suited to the MUG-VOS dataset, we also tested the MUG-VOS dataset with other existing models~\cite{zhang2023personalize, rajivc2023segment, cheng2022xmem, cheng2023tracking}.
Specifically, since our MMPM leverages SAM's knowledge for video segmentation, we implemented a SAM-based baseline model that retrieves the mask proposal with the highest IoU score relative to the previous frame. Masks were initialized by providing grid points to SAM on every frame, and mask tracks were subsequently generated by sequentially selecting the mask with the highest IoU score.

Table~\ref{Table:visa} shows the quantitative results on MUG-VOS dataset.
For the SAM-based baseline, ``Grid" refers to the number of points per side when sampling points from the grid. The baseline demonstrated poor performance because it fails to generate the current frame's mask without relying on prior information from the previous frame. Additionally, solely relying on IoU for track connection makes it vulnerable to significant motion changes and occlusions.
While XMem and DEVA have shown great performance on the VOS task, they performed poorly on the MUG-VOS test set. Our experiments not only demonstrate that the MMPM model achieved the highest score compared to other models but also highlight that the MUG-VOS dataset is more challenging compared to existing VOS benchmarks. Previous VOS benchmarks only evaluate performance on salient objects, allowing models to be tuned to find the most prominent objects. However, to be effective in real-world settings, models should be capable of segmenting diverse object regardless of the spatial position of the target. 

Figure~\ref{fig:quality_comparision} shows the qualitative results compared with other methods. We will also provide more qualitative results in the appendix. MMPM demonstrated superior performance both quantitatively and qualitatively.
We show more qualitative results on the Appendix~\ref{sec:add_qual_results}

\begin{table}[t]
\centering 
\begin{tabular}{l|ccc}
      \toprule
      Methods & $\mathcal{J\&F}$ & $\mathcal{J}$ & $\mathcal{F}$ \\
      \midrule
      Not filter & 85.0 & 85.7 & 84.2 \\
      FIFO & 85.9 & 85.8 & 85.9 \\
      Random & 86.0 & 85.9 & 86.0 \\
      + P. first \& last & 86.1 & 86.0 & 86.1 \\
      \bottomrule
    \end{tabular}
\caption{Ablation studies on memory filtering methods.}
\label{tab:arxiv_memory_filtering_method}
\vspace{-10pt}
\end{table}
\begin{table}[t]
\centering 
\begin{tabular}{l|ccc}
      \toprule
      Methods & $\mathcal{J\&F}$ & $\mathcal{J}$ & $\mathcal{F}$ \\
      \toprule
      No memory & 81.8 & 85.3 & 78.2 \\
      + Only sequential & 81.9 & 85.3 & 78.5 \\
      + Only temporal & 85.7 & 85.9 & 85.8 \\
      + Both & 86.1 & 86.0 & 86.1 \\
      \bottomrule
    \end{tabular}
\caption{Ablation studies on types of memory usage.}
\label{tab:arxiv_memory_usage}
\vspace{-10pt}
\end{table}

\subsection{Ablation study}
We conducted ablation studies on the MMPM model using the MUG-VOS test dataset. These studies explore the effects of different memory filtering methods, the number of memory values, memory update intervals, and the characteristics of each memory module.

Table~\ref{tab:arxiv_memory_filtering_method} shows the performance of the MMPM model with different memory filtering methods. Firstly, no filtering is applied (i.e. Not filter) when memory reaches the limit $T_{max}$ and the memory retains the first inserted values for the entire video. Secondly, to append the latest values to memory, we implemented the First-In-First-Out (FIFO) method. Performance gain from FIFO method enables MMPM model to generate segmentation masks with the latest information enables the model to find corresponding objects efficiently. Thirdly, randomly filtering values from the memory allows the model to access diverse memories with random interval. Lastly, applying a constraint to the random filtering method to preserve the first and last frame values (i.e. P. first \& last), enhanced model performance compared to any other methods. Since, in VOS tasks, the initial mask from the first frame best represents the target object, and providing the latest mask to the MMPM model has a similar effect to the initial mask.

Table~\ref{tab:arxiv_memory_usage} shows the performance comparison between different memory usage types. To demonstrate the advantage of memory usage, we experimented with the MMPM model without memory values, where the mask is generated by referring only to the previous mask, indicated as `No memory'. Adding sequential memory, which is updated from selected frame, improved performance. The low-resolution sequential memory helps the model retrieve the target object. When temporal memory is added, the high-resolution information further enhances the model's ability to segment high-quality masks, as temporal memory stores diverse target information (e.g., shape, color, motion) from multiple frames.

\begin{table}[t]
\centering 
\begin{tabular}{l|ccc}
      \toprule
      \textit{r} & $\mathcal{J\&F}$ & $\mathcal{J}$ & $\mathcal{F}$ \\
      \toprule
      1 & 66.6 & 50.4 & 82.8 \\
      3 & 68.3 & 52.2 & 84.4 \\
      5 & 86.1 & 86.0 & 86.1 \\
      \bottomrule
    \end{tabular}
\caption{Ablation studies on memory update interval.
}
\label{tab:arxiv_memory_interval}
\vspace{-10pt}
\end{table}

\begin{table}[t]
\centering 
\begin{tabular}{l|ccc}
      \toprule
      \textit{N} & $\mathcal{J\&F}$ & $\mathcal{J}$ & $\mathcal{F}$ \\
      \midrule
      5 & 85.7 & 85.8 & 85.5 \\
      10 & 86.1 & 86.0 & 86.1 \\
      15 & 86.1 & 86.0 & 86.1 \\
      \bottomrule
    \end{tabular}
\caption{Ablation studies on number of temporal memory values.
}
\label{tab:arxiv_num_memory_value}
\vspace{-10pt}
\end{table}

Table~\ref{tab:arxiv_memory_interval} shows the performance variation based on the interval period, denoted as \textit{r}, which determines when to update the memory. Increasing the interval between memory updates allows the model to perform faster inference while enabling it to store diverse information in memory. This also demonstrates that our model is robust enough to handle significant appearance changes while effectively mitigating drifting and error accumulation~\cite{oh2019video, yang2021associating}. However, selecting an interval value that is too high can result in the loss of important intermediate information.

Table~\ref{tab:arxiv_num_memory_value} shows the performance variation when implementing different number of memory values in the temporal memory, denoted as \textit{N}. Increasing number of value stored in temporal memory gave MMPM model to look over various memories. Abundant information from past frames enabled model to enhance to image embedding, resulting to MMPM to generate high quality segmentation mask~\cite{duke2021sstvos}. 
Interestingly, MMPM model performance has converged around the number of values 10. 
\section{Conclusion}
In this work, we introduce the MUG-VOS dataset, designed for training and evaluating multi-granularity masks in the video domain. To create this dataset, we developed a data collection pipeline that constructs a large-scale, densely annotated video segmentation dataset, resulting in MUG-VOS. Our dataset includes diverse segmentation masks that are not limited to salient objects but also cover non-salient objects. We also present a video segmentation model named MMPM, which demonstrates superior performance compared to existing models on the MUG-VOS test dataset. We anticipate that the MUG-VOS benchmark will serve as a valuable resource for training and evaluating multi-granularity masks.

\bigskip

\bibliography{aaai25}

\clearpage
\setcounter{section}{0}
\renewcommand{\thesection}{\Alph{section}}
\setcounter{figure}{0}
\renewcommand{\thefigure}{A.\arabic{figure}}
\setcounter{table}{0}
\renewcommand{\thetable}{A.\arabic{table}}

\section*{Appendix}
The following sections present more examination of our results, including a detailed explanation of our methodology, as well as supplementary visualizations of our MUG-VOS and MMPM results. 
Section~\ref{sec:data_collection_pipeline} provide details of data collection pipeline, including annotation tools for constructing video segmentation dataset.
Section~\ref{sec:imp} delves into the specific implementation details of our method, including used losses, hyper-parameter settings, and data augmentations implemented on our training methods.
Section~\ref{sec:sup_experiments} provides additional experiments comparing the performance of MMPM with previous methods.
Section~\ref{sec:MUG_visualization} highlights the quality of our MUG-VOS dataset. Our MUG-VOS dataset not only features a higher density of masks but also maintains high-quality annotations. Furthermore, MUG-VOS includes a diverse range of multi-granularity masks, setting it apart from other existing benchmarks. 
Section~\ref{sec:add_qual_results} provides additional qualitative results on MUG-VOS test set.

\section{A. Details of data collection pipeline}
\setlabel{A}{sec:data_collection_pipeline}
We provide detailed algorithm of MUG-VOS data collection pipeline. See Algorithm \ref{alg:data_engine} for specific details.
\begin{algorithm}
\caption{Data Collection Pipeline Algorithm}
\label{alg:data_engine}
\begin{algorithmic}[1]
\REQUIRE Video $V=\{v_t\vert t\in [1,T]\}$, initial frame $v_1$, SAM model, flow model
\FOR{$t = 1$ to $T$}
    \STATE Obtain foreground masks $M_t^i$ for frame $v_t$
    \IF{$t > 1$}
        \STATE Sample points $\hat{P}_t^i$ from $M_{t-1}^i$ in random
        \STATE Get warped points $P_t^i = \mathrm{warp}(\hat{P}_t^i, F_{t-1\rightarrow t})$
        \STATE Predict candidate masks $\hat{M}_t^i = \mathrm{SAM}(v_t, P_t^i)$
        \STATE Get warped mask $\tilde{M}_{t-1}^i = \mathrm{warp}(M_{t-1}^i, F_{t-1\rightarrow t})$
        \STATE Compute IoU between $\tilde{M}_{t-1}^i$ and $\hat{M}_t^i$
        \STATE Select $M_t^i = \arg\max_{\hat{M}_t^i}\mathrm{IoU}(\tilde{M}_{t-1}^i, \hat{M}_t^i)$
    \ELSE
        \STATE Initialize $M_1^i$ using SAM with grid point prompts
    \ENDIF
\ENDFOR
\end{algorithmic}
\end{algorithm}

We also developed specialized annotation tools to enhance the efficiency of our workflow, as shown in Fig. 5.  Initially, supervisors select multi-granularity masks from the first frame, which include categories such as ``stuff", ``thing", or ``non-salient objects" that need to be tracked, using SAM (Segment Anything Model)~\cite{kirillov2023segment}. Annotators then verify whether the masks generated by the data collection pipeline are consistent with those from the first and previous frames. If the masks are inconsistent, annotators refine the masks by prompting points to SAM. Once a mask track is complete, the results are reviewed by supervisors to ensure the track is both consistent and smooth.

As SAM provides stability and IoU prediction scores, annotators can use these scores to ensure they generated the fine-grained masks. Additionally, besides picking foreground points during the refine process, annotators can also select negative points to obtain the desired masks.

\section{B. Implementation details}\setlabel{B}{sec:imp}
\subsection{B.1 Architecture}
We employ MMPM, which leverages the SAM image encoder pretrained on the SA-1B dataset. The pretrained SAM encoder, based on the Vision Transformer architecture, is initialized with `image segment anything' weights. We use SAM-Base as our visual encoder. Additionally, we introduce a memory architecture comprising two types of memory. These memory types enhance image features by utilizing previous results, enabling the model to effectively segment multi-granularity objects by integrating information from previous frames with current image details.

To efficiently combine information from multiple frames, we implemented an affinity matrix based on a similarity matrix. The similarity matrix is calculated by comparing key elements with the query element, using the same similarity calculation method as in STCN~\cite{cheng2021rethinking} and XMem~\cite{cheng2022xmem}. The affinity matrix is then obtained by applying the softmax function to the similarity matrix.

\subsection{B.2 Training details}
In this subsection, we detail the training procedures for MMPM, XMem, and DEVA. To ensure a fair comparison, we trained XMem and DEVA on the MUG-VOS training set alongside MMPM.
For MMPM training, we divided the process into two stages. First, we trained MMPM on a static image dataset, applying independent augmentations to simulate video-like sequences. This step helps the model learn to handle static images as if they were part of a video. In the second stage, we trained MMPM on the MUG-VOS training set. We used the AdamW~\cite{loshchilov2017decoupled} optimizer with a learning rate of 1e-5. The model was trained for 200K iterations with a batch size of 3, followed by 250K iterations with a batch size of 4.

To input an image into the SAM encoder, we selected the longest axis of the image and resized it to 1024 pixels, scaling the other axis proportionally to maintain the original aspect ratio.
To increase the complexity during training, we augmented the MUG-VOS dataset with random cropping, horizontal flipping, and affine transformations. Additionally, given that we are working with a video dataset, we randomly shuffled the image order to improve the model's learning of video sequences.

For XMem and DEVA, we used the same dataset that was employed for MMPM training, adhering to the training procedures outlined in their respective papers. Specifically, we trained DEVA on a static image dataset with a batch size of 32 for 40K iterations. For XMem, we utilized the stage 0 model provided by the original repository.

After training on the static dataset, we directly trained both models on the MUG-VOS training set. XMem was trained for an additional 11K iterations with batch size 16, while the DEVA model was trained for 7.5K iterations with batch size 16.

\section{C. Experimental results}
\setlabel{C}{sec:sup_experiments}

\begin{table}[t]
\centering
\resizebox{\linewidth}{!}{ 
\begin{tabular}{lccccc}
\toprule
DAVIS-2017 & $\mathcal{J}\&\mathcal{F}$ & $\mathcal{J}$  & $\mathcal{F}$ \\
\midrule
XMem~\cite{cheng2022xmem}  & 63.0 & 59.7 & 66.4 \\
DEVA~\cite{cheng2023tracking}  & 64.1 & 60.8 & 67.4 \\
MMPM  & \underline{69.1} & \underline{66.0} & \underline{72.1} \\
\bottomrule
\end{tabular}
}
\caption{The quantitative evaluation of video object segmentation on DAVIS-2017~\cite{pont20172017} validation set. Note that all the models are trained on the MUG-VOS train set.}
\vspace{-10pt}
\label{Table:vos}
\end{table}

We evaluated our method on the DAVIS-2017~\cite{pont20172017} dataset. To ensure a fair comparison, all trainable methods listed in Table~\ref{Table:visa}, including MMPM, were trained uniformly on the MUG-VOS Train set and utilized the same mask prompt to standardize input and output formats. Furthermore, DEVA, XMem, and MMPM were exclusively trained on the MUG-VOS Train set and subsequently evaluated on DAVIS-2017. Our results are presented in Table~\ref{Table:vos}.

\section{D. MUG-VOS visualization}
\setlabel{D}{sec:MUG_visualization}
We present the qualitative results of the MUG-VOS test dataset, which includes 0.663 density, 59k masks, 887 mask tracks, 29.6 masks per frame, and 1,999 annotated frames, as shown from Fig.~\ref{fig:mug_test_1} to Fig.~\ref{fig:mug_test_2}. Additionally, we provide a qualitative comparison between the ground truth masks of existing benchmarks and the masks generated by our data collection pipeline, illustrated from Fig.~\ref{fig:dataset_comp_0} to Fig.\ref{fig:dataset_comp_2}. Our data collection pipeline allows us to construct a dataset annotated with masks of more varied granularity than those found in existing benchmarks. Further visualizations of the MUG-VOS dataset, organized by the number of mask tracks, are presented from Fig.~\ref{fig:mug_qual_0} to Fig.~\ref{fig:mug_qual_1}.

\section{E. Additional qualitative results}
\setlabel{E}{sec:add_qual_results}
We provide additional qualitative results of MMPM on our proposed MUG-VOS dataset from Fig.\ref{fig:pred_comp_0} to Fig.\ref{fig:pred_comp_1}. Our MMPM demonstrates superior results compared to other methods. As shown in the figures, other methods, except for MMPM, either made tracking errors or generated unstable masks. However, MMPM confidently produced fine-grained segmentation masks, highlighting its ability to accurately track and segment multi-granularity objects.

\begin{figure*}[t]
    \includegraphics[width=\linewidth]{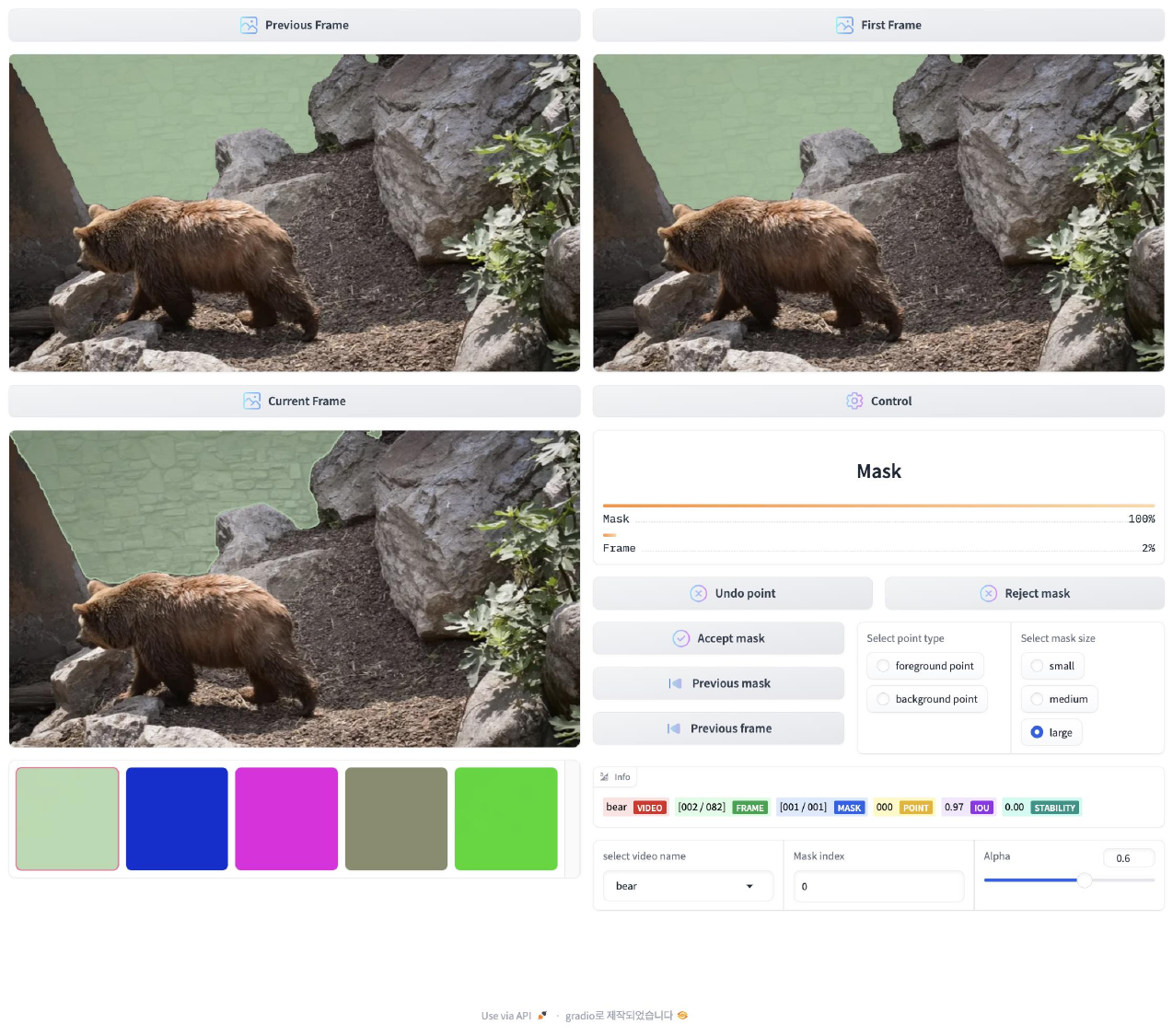}
    \caption{\textbf{MUG-VOS annotation tool.} We developed mask annotation tool for curating video segmentation dataset. }
    \label{fig:anno_tool}
\end{figure*}
\begin{figure*}[t]
    \includegraphics[width=\linewidth]{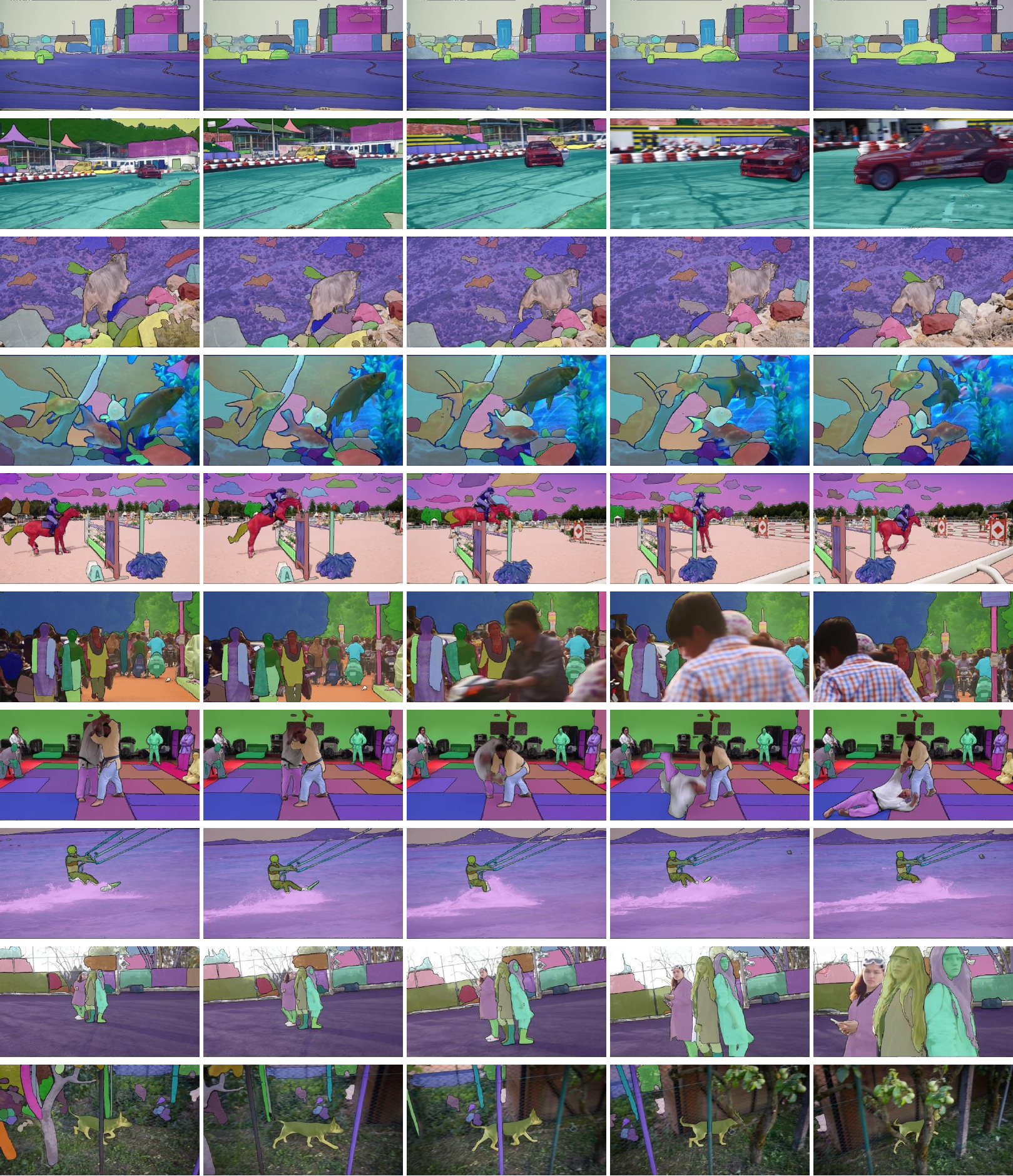}
    \caption{\textbf{MUG-VOS test dataset.} To mitigate the accumulation of errors within the automated process, annotators were directed to either approve or reject the mask tracks generated by the data collection pipeline. In instances where errors were detected, the annotators performed frame-level refinements of the masks utilizing the Segment Anything Model.}
    \label{fig:mug_test_1}
\end{figure*}
\begin{figure*}[t]
    \includegraphics[width=\linewidth]{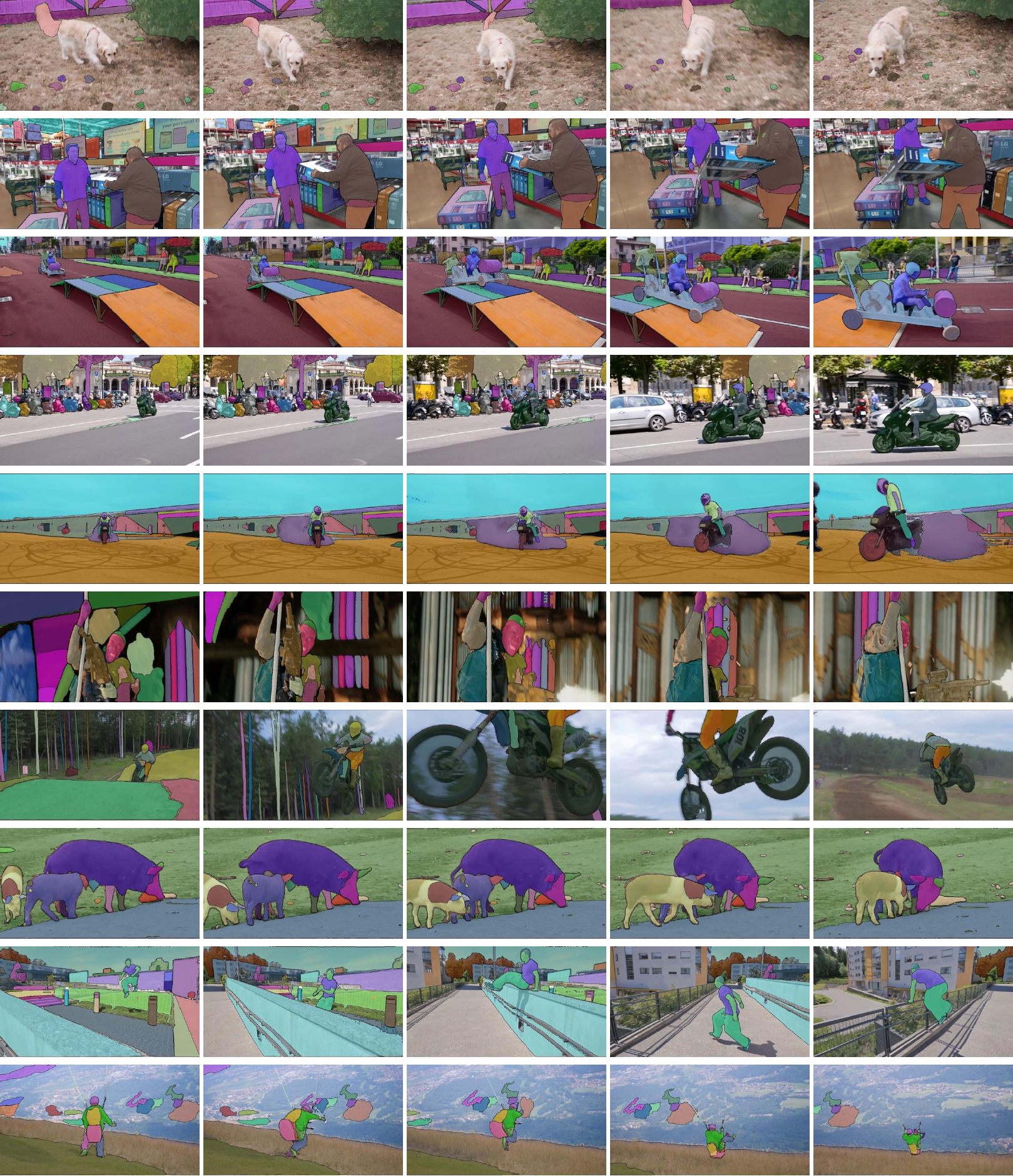}
    \caption{\textbf{MUG-VOS test dataset.} To mitigate the accumulation of errors within the automated process, annotators were directed to either approve or reject the mask tracks generated by the data collection pipeline. In instances where errors were detected, the annotators performed frame-level refinements of the masks utilizing the Segment Anything Model.}
     \label{fig:mug_test_2}
\end{figure*}
\begin{figure*}[t]
    \centerline{\includegraphics[height=\textheight]{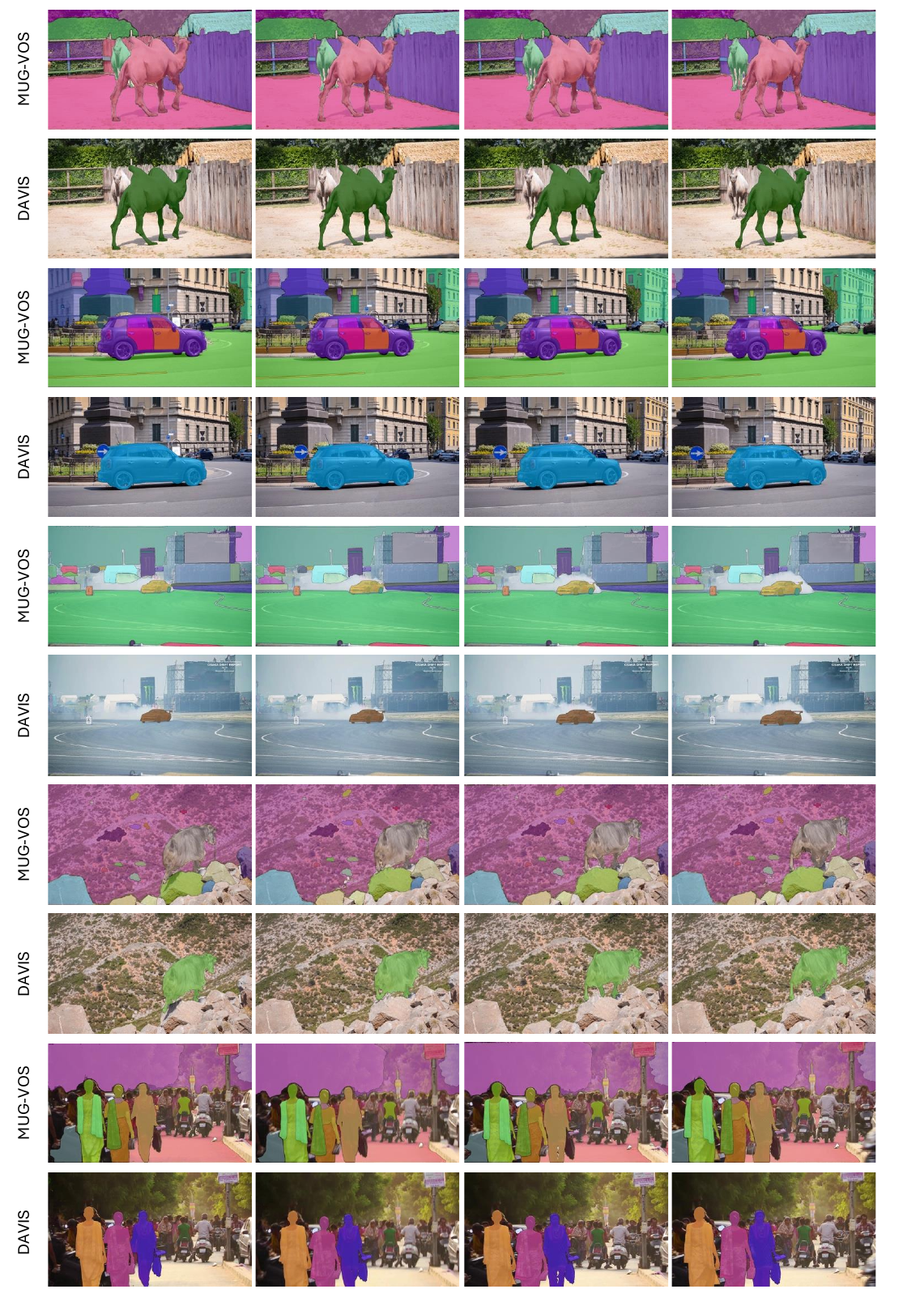}}
    \caption{\textbf{Mask comparison.} We qualitatively compare mask generation quality of \textbf{our data collection pipeline} with ground-truth from DAVIS~\cite{pont20172017}, Youtube-VOS~\cite{xu2018youtube}, and UVO~\cite{wang2021unidentified}.}
    \label{fig:dataset_comp_0}
\end{figure*}
\begin{figure*}[t]
    \centerline{\includegraphics[height=\textheight]{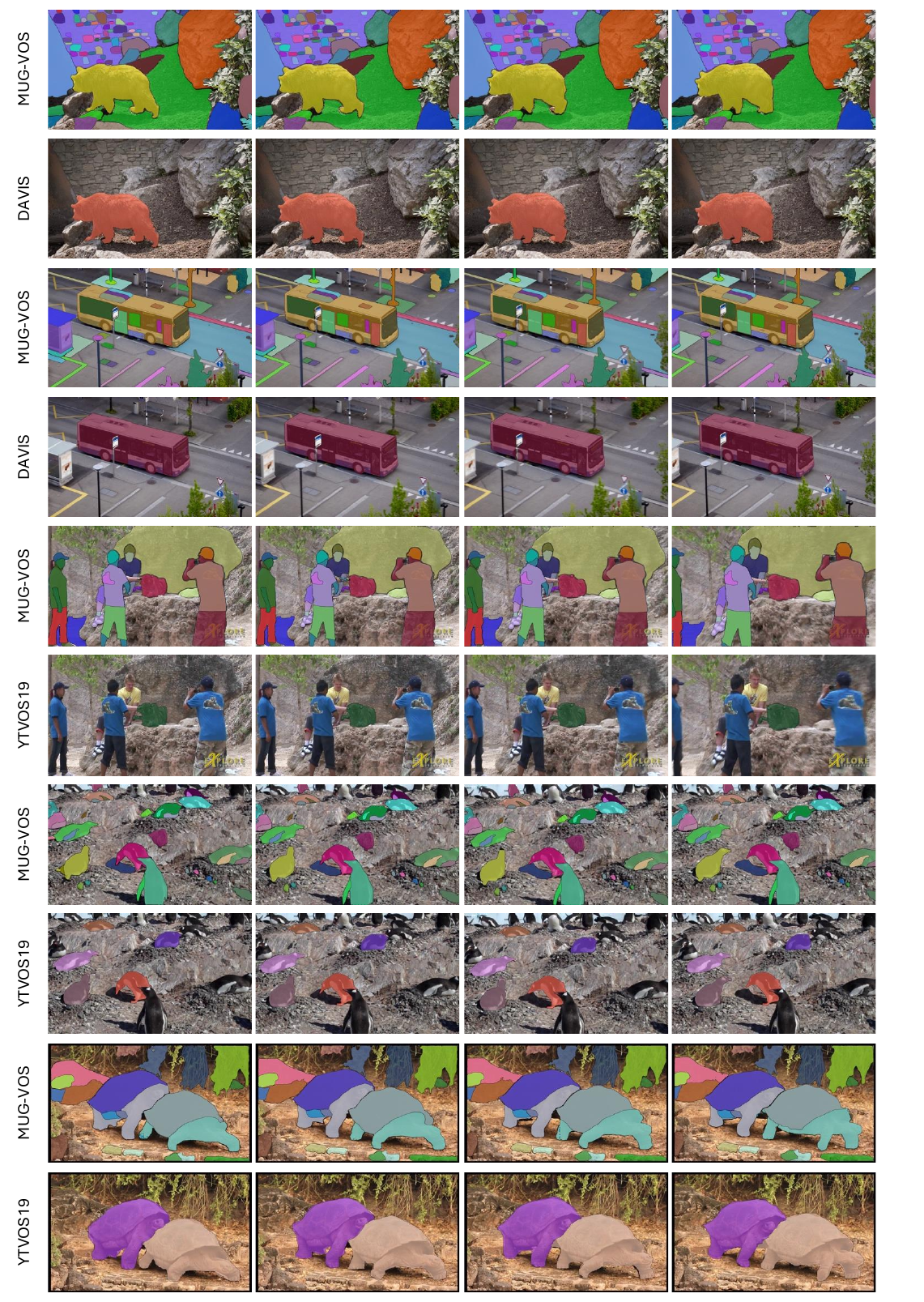}}
    \caption{\textbf{Mask comparison.} We qualitatively compare mask generation quality of \textbf{our data collection pipeline} with ground-truth from DAVIS~\cite{pont20172017}, Youtube-VOS~\cite{xu2018youtube}, and UVO~\cite{wang2021unidentified}.}
    \label{fig:dataset_comp_1}
\end{figure*}
\begin{figure*}[t]
    \centerline{\includegraphics[height=\textheight]{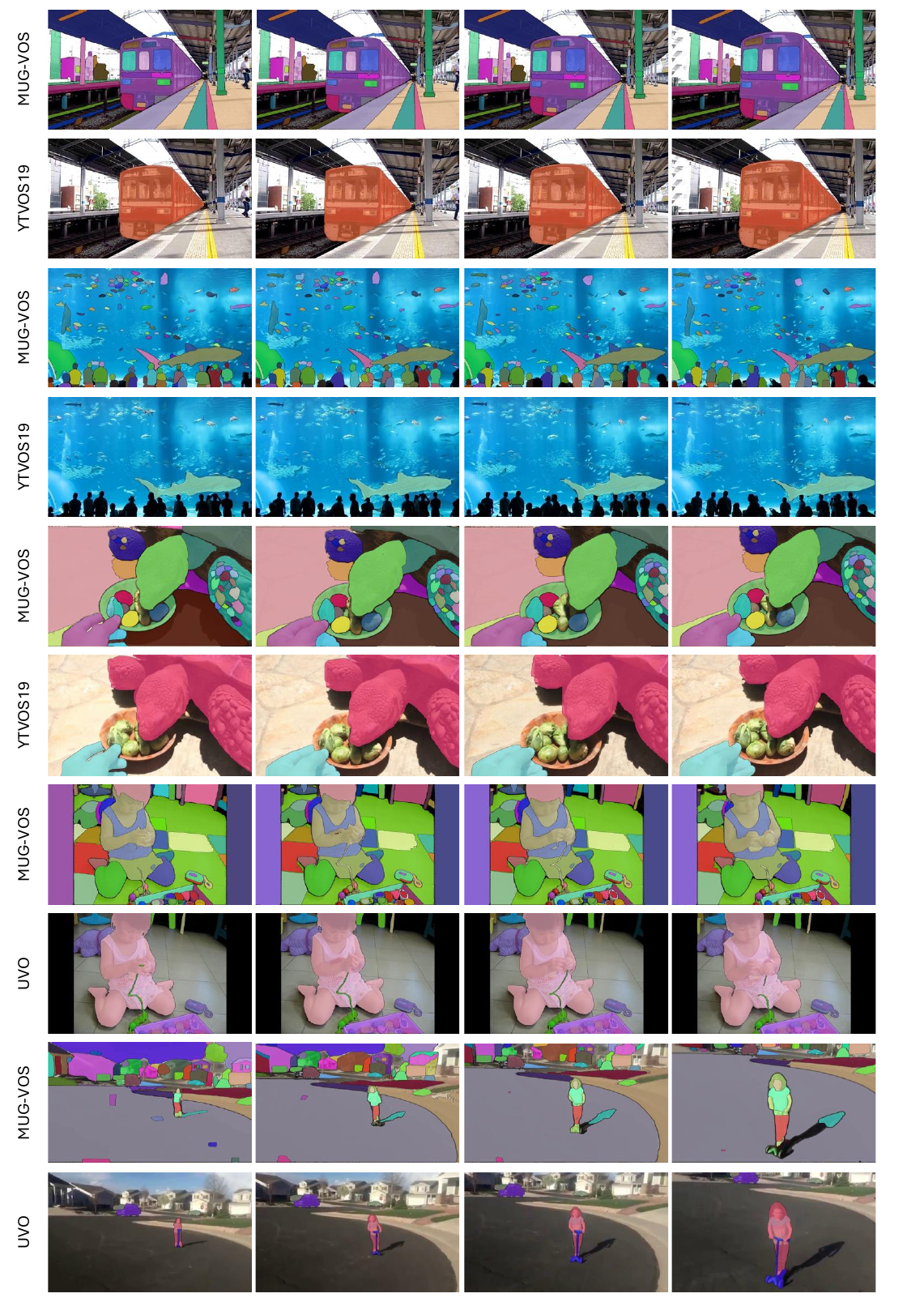}}
    \caption{\textbf{Mask comparison.} We qualitatively compare mask generation quality of \textbf{our data collection pipeline} with ground-truth from DAVIS~\cite{pont20172017}, Youtube-VOS~\cite{xu2018youtube}, and UVO~\cite{wang2021unidentified}.}
     \label{fig:dataset_comp_2}
\end{figure*}

\begin{figure*}[t]
    \centerline{\includegraphics[height=\textheight]{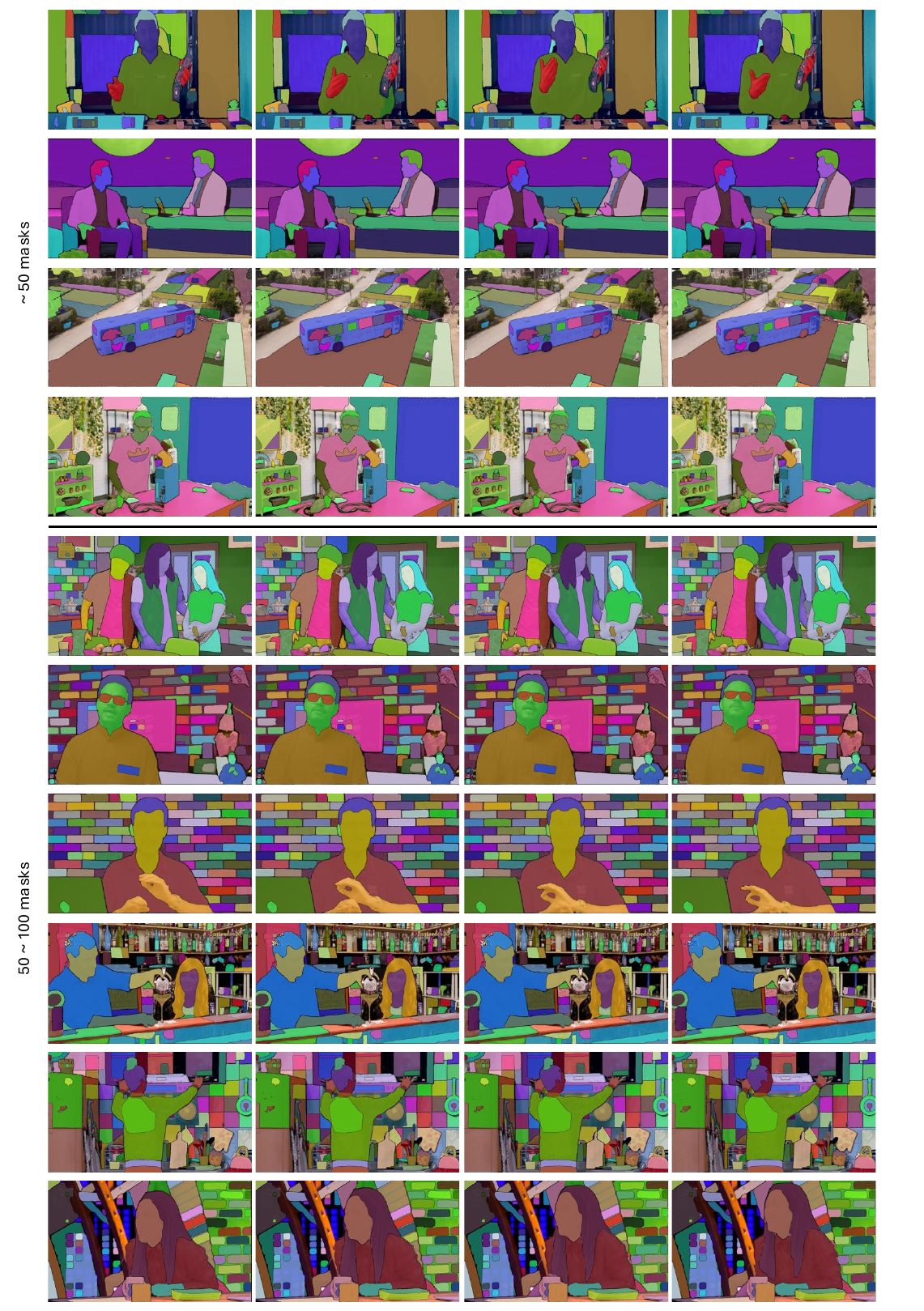}}
    \caption{\textbf{Example image sequences with overlaid masks from our MUG-VOS dataset.} These masks are generated by our data collection pipeline \textit{fully automatically}. We group videos by number of masks per frame for visualization.}
    \label{fig:mug_qual_0}
\end{figure*}

\begin{figure*}[t]
    \centerline{\includegraphics[height=\textheight]{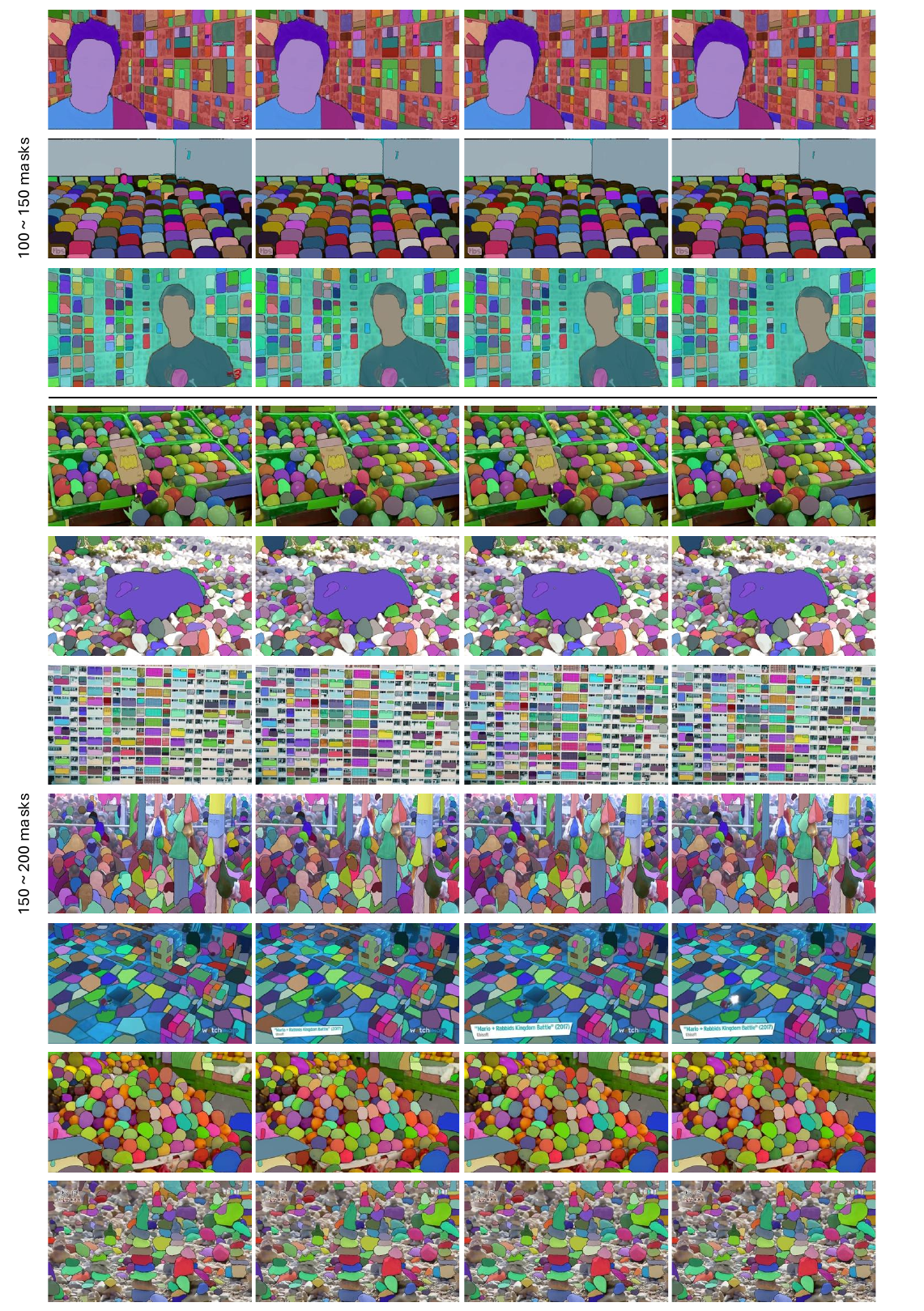}}
    \caption{\textbf{Example image sequences with overlaid masks from our MUG-VOS dataset.} These masks are generated by our data collection pipeline \textit{fully automatically}. We group videos by number of masks per frame for visualization.}
    \label{fig:mug_qual_1}
\end{figure*}
\begin{figure*}[t]
    \includegraphics[width=\linewidth]{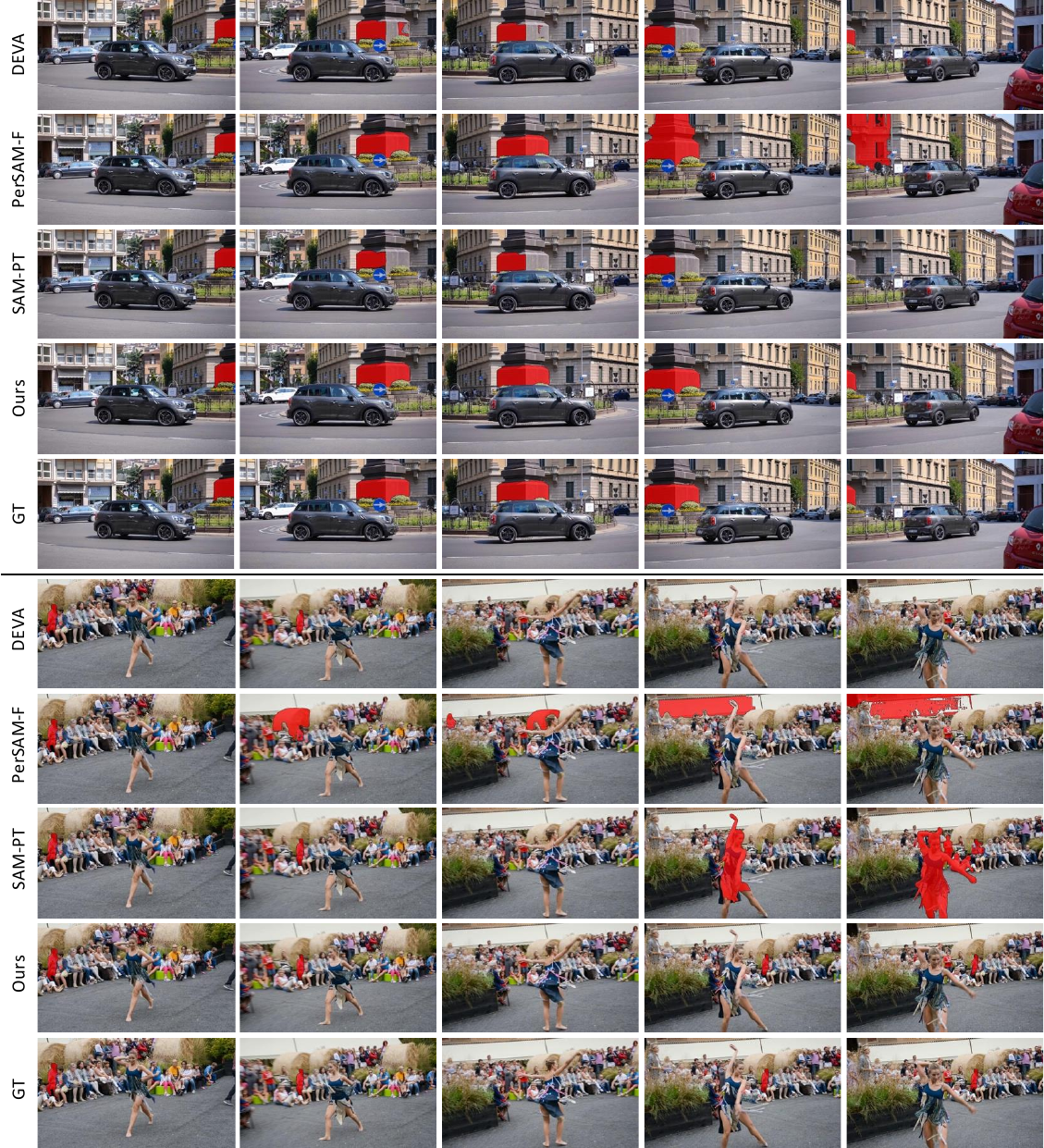}
    \caption{\textbf{Qualitative comparison.} We qualitatively compare mask generation quality of \textbf{our data collection pipeline} with ground-truth from DAVIS~\cite{pont20172017}, Youtube-VOS~\cite{xu2018youtube}, and UVO~\cite{wang2021unidentified}.}
    \label{fig:pred_comp_0}
\end{figure*}
\begin{figure*}[t]
    \includegraphics[width=\linewidth]{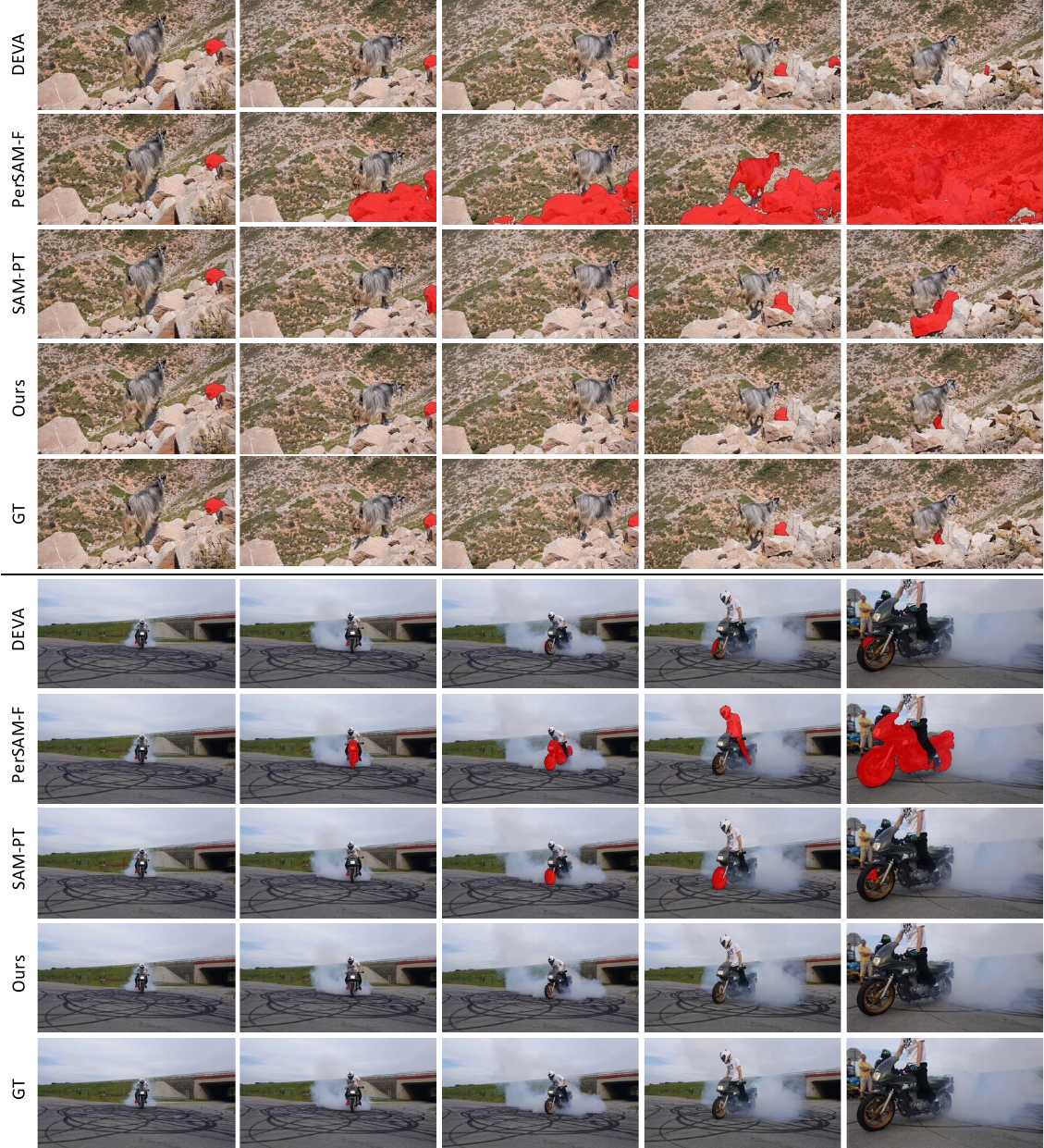}
    \caption{\textbf{Qualitative comparison.} We qualitatively compare mask generation quality of \textbf{our data collection pipeline} with ground-truth from DAVIS~\cite{pont20172017}, Youtube-VOS~\cite{xu2018youtube}, and UVO~\cite{wang2021unidentified}.}
    \label{fig:pred_comp_1}
\end{figure*}

\end{document}